\documentclass[letterpaper]{article}
\usepackage{aaai}
\usepackage{times}
\usepackage{helvet}
\usepackage{courier}
\usepackage{amssymb}
\usepackage{amsmath}
\usepackage{graphicx}
\usepackage{algorithmic,algorithm}
\usepackage{subcaption}
\usepackage{listings}
\usepackage{hyperref}
\usepackage{multirow}
\usepackage{array}

\newcolumntype{C}[1]{>{\centering}m{#1}}
\newcommand{\paratitle}[1]{\vspace{1.5ex}\noindent \textbf{#1}}
\newcommand*\samethanks[1][\value{footnote}]{\footnotemark[#1]}

\usepackage{bm}
\title{RLCard: A toolkit for Reinforcement Learning in Card Games}

\begin{document}
%
\title{RLCard: A Toolkit for Reinforcement Learning in Card Games}
\author{Daochen Zha$^1$, Kwei-Herng Lai$^1$, Yuanpu Cao$^1$\thanks{Authors contribute during the visit at Texas A\&M University.}, Songyi Huang$^2$, Ruzhe Wei$^1$\samethanks, Junyu Guo$^1$\samethanks, Xia Hu$^1$\\
$^1$ Department of Computer Science and Engineering, Texas A\&M University, College Station, USA \\
$^2$ Simon Fraser University, BC, Canada \\
\{daochen.zha, khlai037\}@tamu.edu, yuanpucao@gmail.com, songyih@sfu.ca,\\
ruzhe.wei@outlook.com, \{guojunyu, xiahu\}@tamu.edu
}
\maketitle

\maketitle
\begin{abstract}
We present RLCard, an open-source toolkit for reinforcement learning research in card games. It supports various card environments with easy-to-use interfaces, including Blackjack, Leduc Hold'em, Texas Hold'em, UNO, Dou Dizhu and Mahjong. The goal of RLCard is to bridge reinforcement learning and imperfect information games, and push forward the research of reinforcement learning in domains with multiple agents, large state and action space, and sparse reward. In this paper, we provide an overview of the key components in RLCard, a discussion of the design principles, a brief introduction of the interfaces, and comprehensive evaluations of the environments. The codes and documents are available at \url{https://github.com/datamllab/rlcard}.
\end{abstract}

\section{Introduction}
\label{sec:1}
Reinforcement learning~(RL) is a promising paradigm in Artificial Intelligence for learning goal-oriented tasks. Through interactions with the environments, reinforcement learning agents learn to make decisions at each state in a trial-and-error fashion. With neural networks as function approximators, deep reinforcement learning has recently achieved breakthroughs in various domains: Atari games~\cite{mnih2015human}, Go game~\cite{silver2017mastering}, continuous control~\cite{lillicrap2015continuous}, and neural architecture search~\cite{zoph2016neural}, just to name a few. Out of these achievements, however, reinforcement learning is still immature and unstable in applications with multiple agents, large decision space or sparse reward.


In this paper, we introduce various styles of card environments for reinforcement learning research. Card games are ideal testbeds with several challenges. \textbf{First}, card games are played by multiple agents who must learn to compete or collaborate with each other. For example, in Dou Dizhu, peasants need to work together to fight against the landlord in order to win the game. \textbf{Second}, card games have huge state space. For instance, the number of states in UNO can reach $10^{163}$. The cards of each player are hidden from the other players. A player not only needs to consider her own hand, but also has to reason about the other players' cards from the signals of their actions. \textbf{Third}, card games may have large action space. For example, the possible number of actions in Dou Dizhu can reach $10^4$ with an explosion of card combinations. \textbf{Last}, card games may suffer from sparse reward. For example, in Mahjong, winning hands are scarce. We observe one winning hand every five hundreds of games if playing randomly. Moreover, card games are easy to understand with huge popularity. Games such as Texas Hold'em, UNO and Dou Dizhu are played by hundreds of millions of people. We usually do not need to spend efforts on learning the rules before we can dive into algorithm development.


\begin{figure}[t]
    \centering
    \includegraphics[width=6cm]{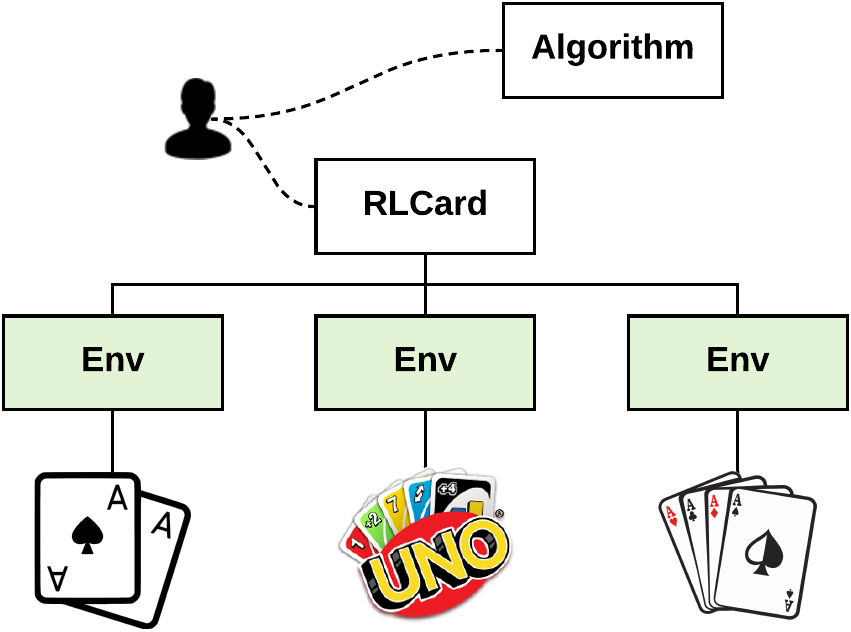}
    \caption{An overview of RLCard. It supports various styles of card games, such as betting games, Chinese Poker, and boarding games, wrapped by easy-to-use interfaces.}
    \vspace{-17pt}
    \label{fig:1}
\end{figure}

\begin{table*}[t]
    \centering
    \begin{tabular}{l|c|c|c}
        \hline  
         \textbf{Environment} & \textbf{InfoSet Number} & \textbf{Avg. InfoSet Size} & \textbf{Action Size}\\
         \hline 
         Blackjack & $10^3$ & $10^1$ &$10^0$\\
         \hline
         Leduc Hold'em & $10^{2}$ & $10^2$ &$10^0$\\
         \hline
         Limit Texas Hold'em & $10^{14}$ & $10^3$ &$10^0$\\
         \hline
         Dou Dizhu & $10^{53} \sim 10^{83}$ & $10^{23}$ &$10^4$\\
         \hline
         Mahjong & $10^{121}$ & $10^{48}$ &$10^2$\\
         \hline
         No-limit Texas Hold'em & $10^{162}$ & $10^3$ &$10^4$\\
         \hline
         UNO & $10^{163}$ & $10^{10}$ &$10^1$\\
         \hline  
    \end{tabular}
    \caption{A summary of the games in RLCard. \textbf{InfoSet Number:} the number of the information sets; \textbf{Avg. InforSet Size:} the average number of states in a single information set; \textbf{Action Size:} the size of the action space (without abstraction). Note that for some games, we can only provide a range of the complexity estimation. For example, Dou Dizhu allows a large number of legal combinations of cards, which makes it challenging to estimate the size of the state space.}
    \label{tab:summary}
        \vspace{-12pt}
\end{table*}

To develop card environments with easy-to-use interfaces is a challenging task. First, the interfaces must be accessible to RL researchers who may or may not have a game theory background. In the extensive form games, the player will not observe her next state immediately after taking an action. The next state is exposed to the player only after all other players have chosen their actions. This makes it difficult to design environment interfaces. Second, the environments need to be configurable. The state representation, action abstraction, reward design, or even the game rules should be easily adjusted for research purposes.

We present RLCard, an opensource toolkit designed for reinforcement learning in card games. It supports various card environments, as summarized in Table~\ref{tab:summary}. The interfaces are straightforward for reinforcement learning. The transitions of each player and collected and well organized after a complete game in the multi-agent setting. We also provide a single-agent interface, where the other players are simulated using pre-trained models. The state and action encoding can be easily configured. The games are implemented under the same structure with clear logic. The evaluation tools are provided to measure the performance by winning rates of tournaments. Future versions will extend the toolkit to include more environments. The goal of RLCard is to bridge reinforcement learning and imperfect information games, and push forward the research of reinforcement learning in domains with multiple agents, large state space, large action space, and sparse reward.

\vspace{-4pt}
\section{Overview}
\label{sec:2}
In this section, we give an overview of RLCard, and introduce the interfaces. More introductions can be found in Appendix. Figure~1 shows an overview of RLCard. Each game is wrapped by an environment class with easy-to-use interfaces. With RLCard, we can focus on algorithm development instead of engineering efforts on games. When developing the toolkit, we adopt the following design principles:
\begin{itemize}
	\item \textbf{Reproducible.} Results on the environments can be reproduced. The same result should be obtained with the same random seed in different runs.
	\item \textbf{Accessible.} Experiences are collected and well organized after each game with straightforward interfaces. State representation, action encoding, reward design, or even the game rules, can be conveniently configured.
	\item \textbf{Scalable.} New card environments can be added conveniently into the toolkit with the above design principles. We try to minimize the dependencies in the toolkit so that the codes can be easily maintained.

\end{itemize}

\vspace{-4pt}
\subsection{Available Environments}
The toolkit provides various styles of card games that are popular among hundreds of millions of people, including betting games, Chinese Poker, and some boarding games. Table~\ref{tab:summary} summarizes the card games in RLCard and estimates the complexity of each game. The game size can be measured by the number of information sets, which are the observed states from the view of one player. The average size of the information set is defined as the average number of possible game states in each information set. For example, given the observation from the view of one player in Texas Hold'em, the other players could have many possible hands. Each possible hand corresponds to one game state in this information set. The size of the action space is also provided since large action space will greatly increase the difficulty.

Since most of games have very large state space, it is challenging to immediately solve these human-size games. Thus, we have also implemented some smaller versions of some large games. For example, RLCard also implements a smaller version of Dou Dizhu, where we only keep cards 8, 9, 10, J, Q, K, and A. This variant keeps key features of Dou Dizhu but with much smaller state space.

\vspace{-4pt}


\subsection{Basic Interface}

We provide a \texttt{run} function for quickly getting started. It directly generates payoffs and game data, which are organized as transitions, i.e., (state, action, reward, next\_state, done). This interface is designed for algorithms that do not need to traverse the game tree. An example of running Dou Dizhu with three random agents is as follows: 

\lstset{language=Python}
\lstset{frame=lines}
\lstset{label={lst:code_direct}}
\lstset{basicstyle=\footnotesize}
\begin{lstlisting}
import rlcard
import RandomAgent

# Initialize the environment
env = rlcard.make('doudizhu')

# Initialize random agents
agent = RandomAgent()
env.set_agents([agent, agent, agent])

while True:
	# Generate data from the environment
	trajectories, payoffs = env.run()
	# Train agent here
\end{lstlisting}
For sampling based algorithms that do not require traversing backward in the game tree~\cite{heinrich2015fictitious,heinrich2016deep,lanctot2017unified}, the basic interface could be preferred since we do not need to care about the details of the traversing.

\begin{figure*}
  \centering
  \begin{subfigure}[b]{0.24\textwidth}
    \includegraphics[width=\textwidth]{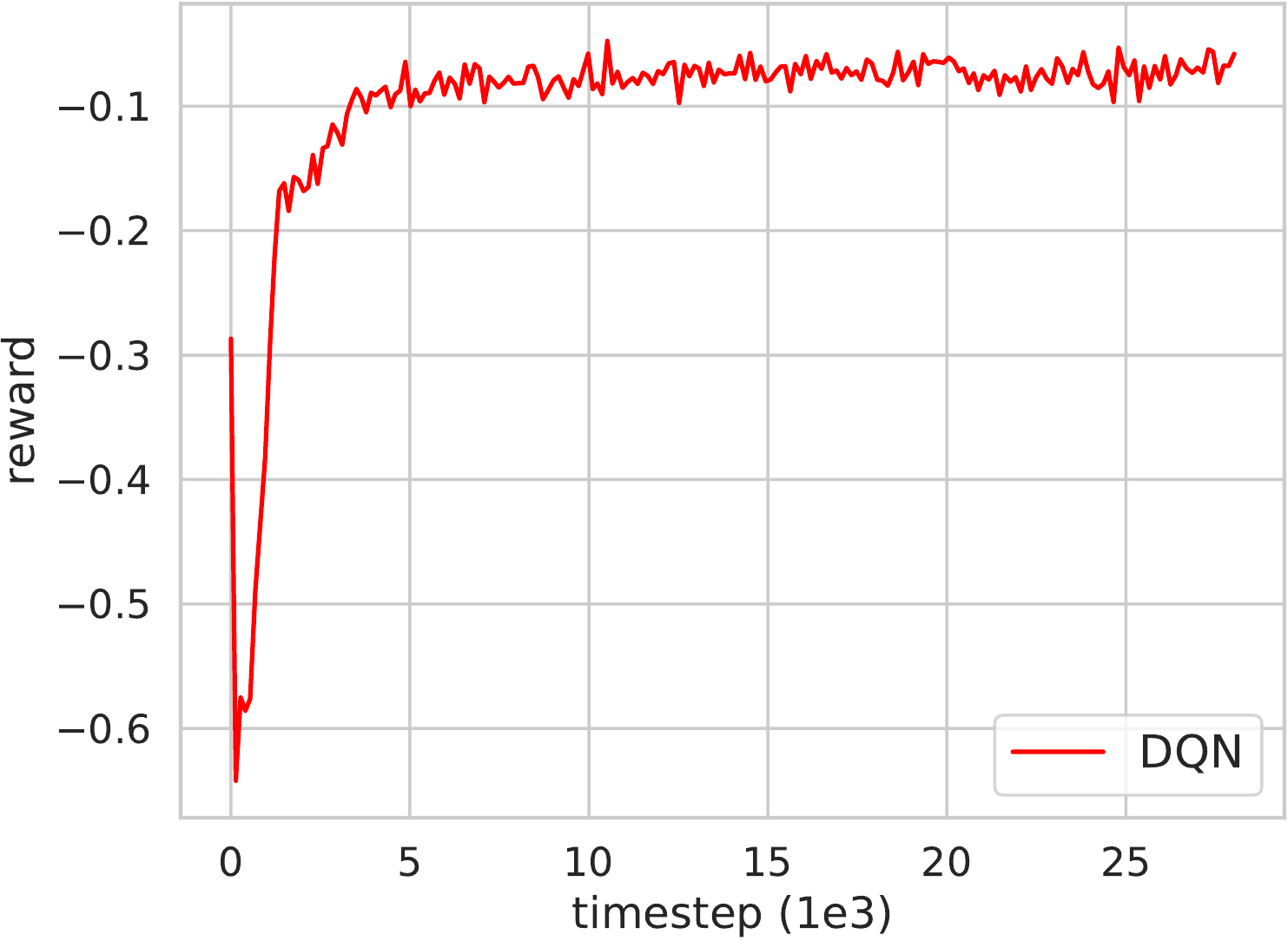}
    \caption{Blackjack}
    \vspace{15pt}
  \end{subfigure}%
  \begin{subfigure}[b]{0.24\textwidth}
    \includegraphics[width=\textwidth]{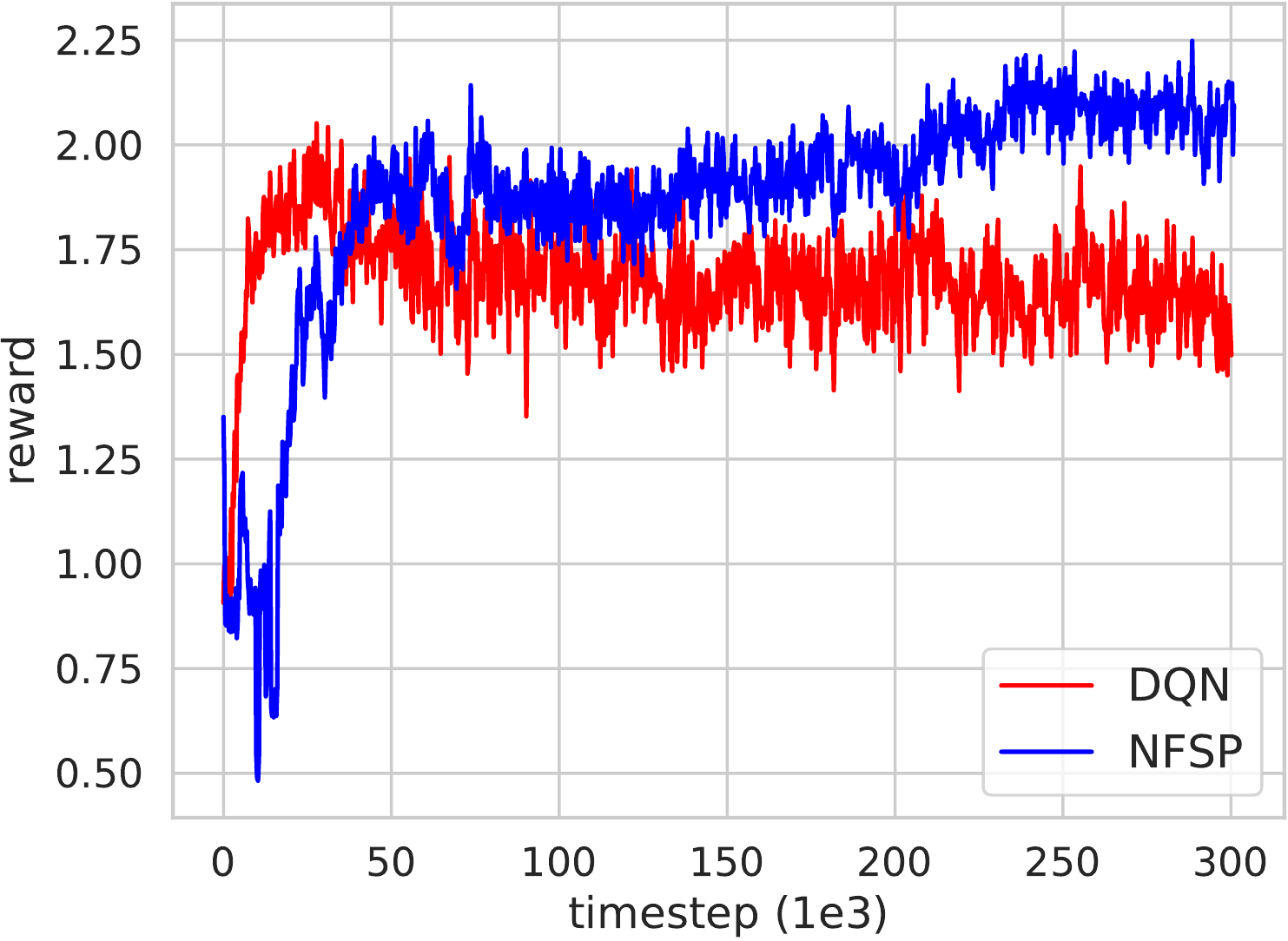}
    \caption{Leduc Hold'em}
    \vspace{15pt}
  \end{subfigure}%
  \begin{subfigure}[b]{0.24\textwidth}
    \includegraphics[width=\textwidth]{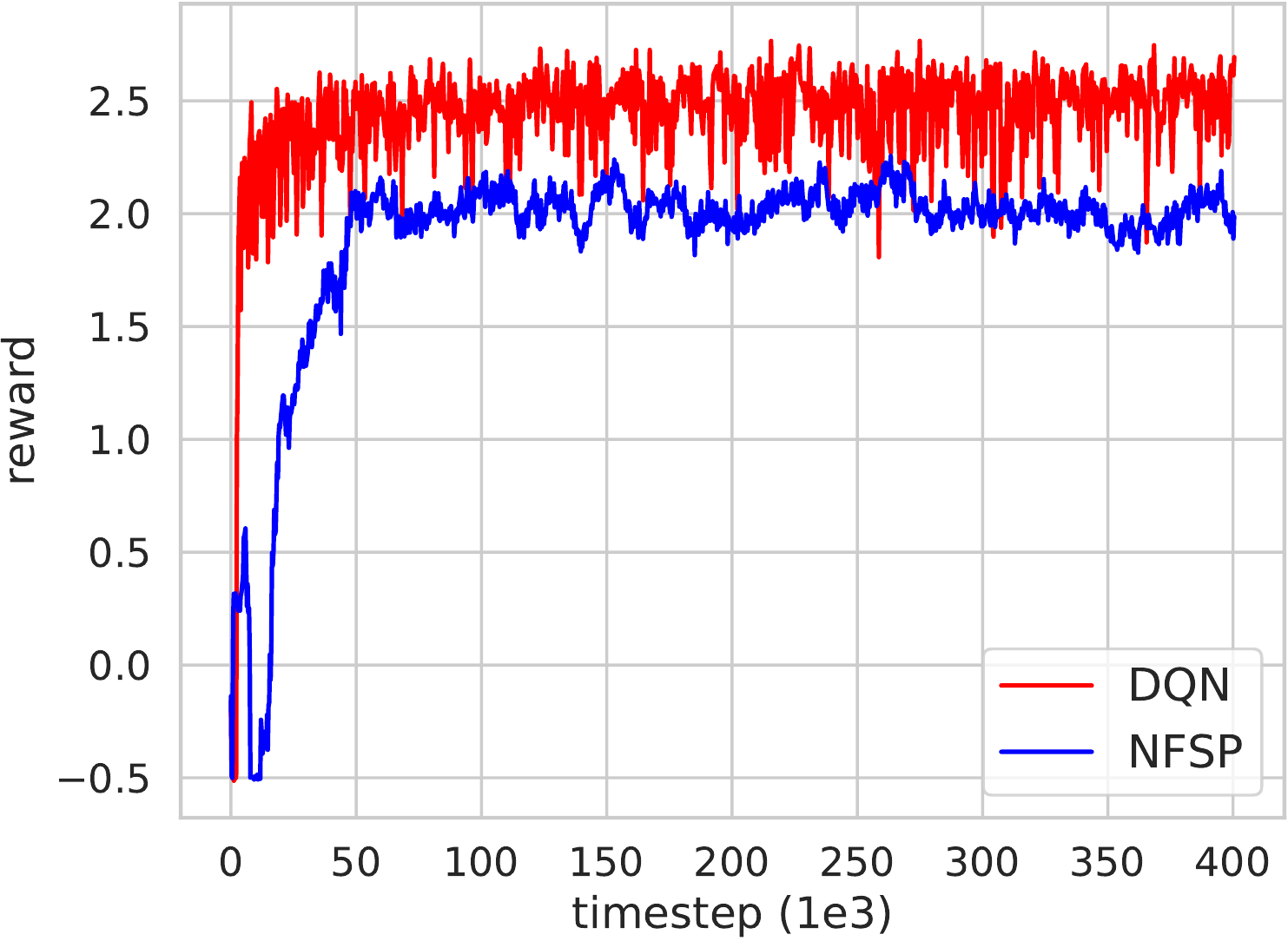}
    \caption{Limit Texas Hold'em}
    \vspace{15pt}
  \end{subfigure}%
  \begin{subfigure}[b]{0.24\textwidth}
    \includegraphics[width=\textwidth]{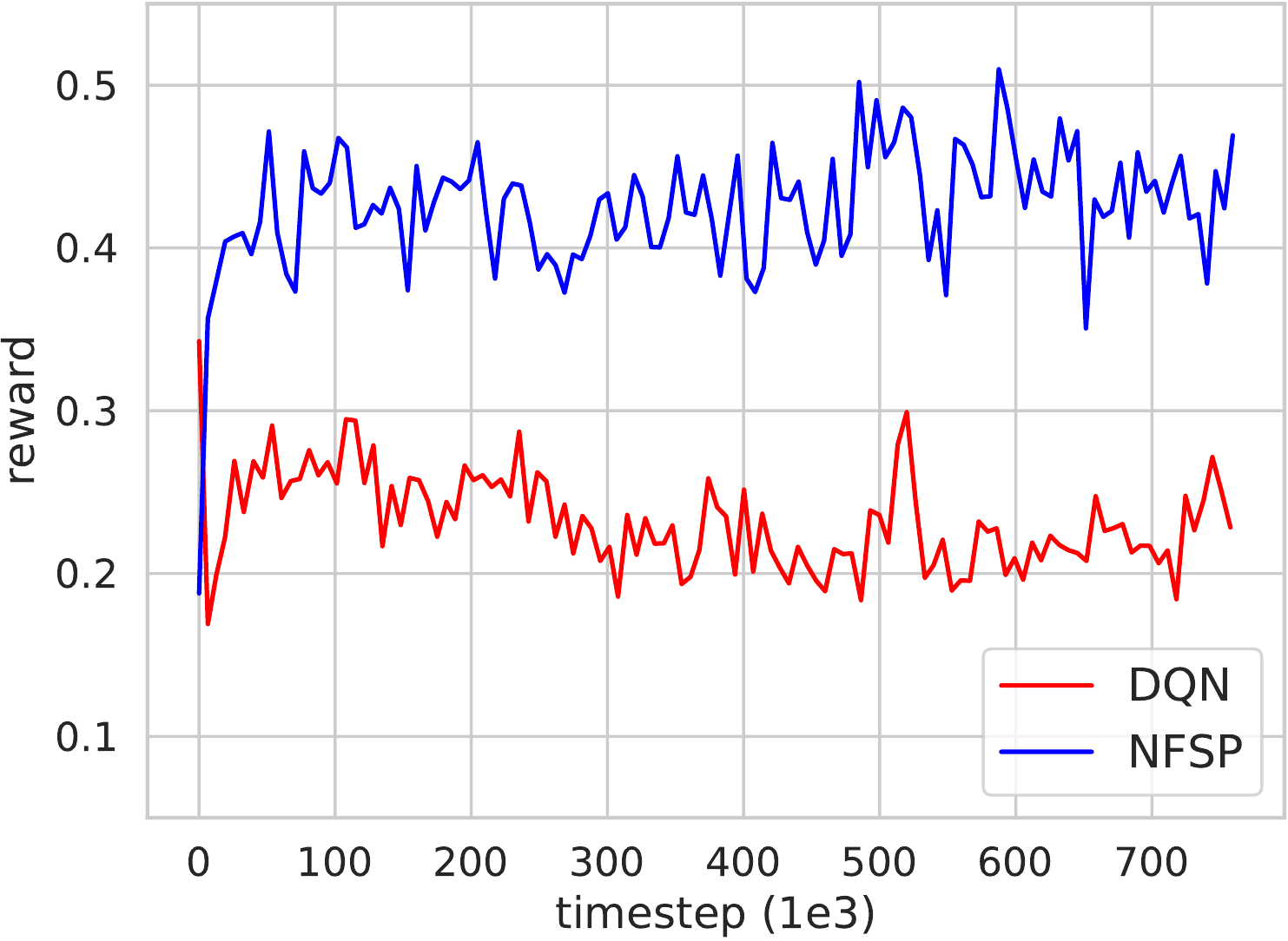}
    \caption{Dou Dizhu}
    \vspace{15pt}
  \end{subfigure}%
  
  \begin{subfigure}[b]{0.24\textwidth}
    \includegraphics[width=\textwidth]{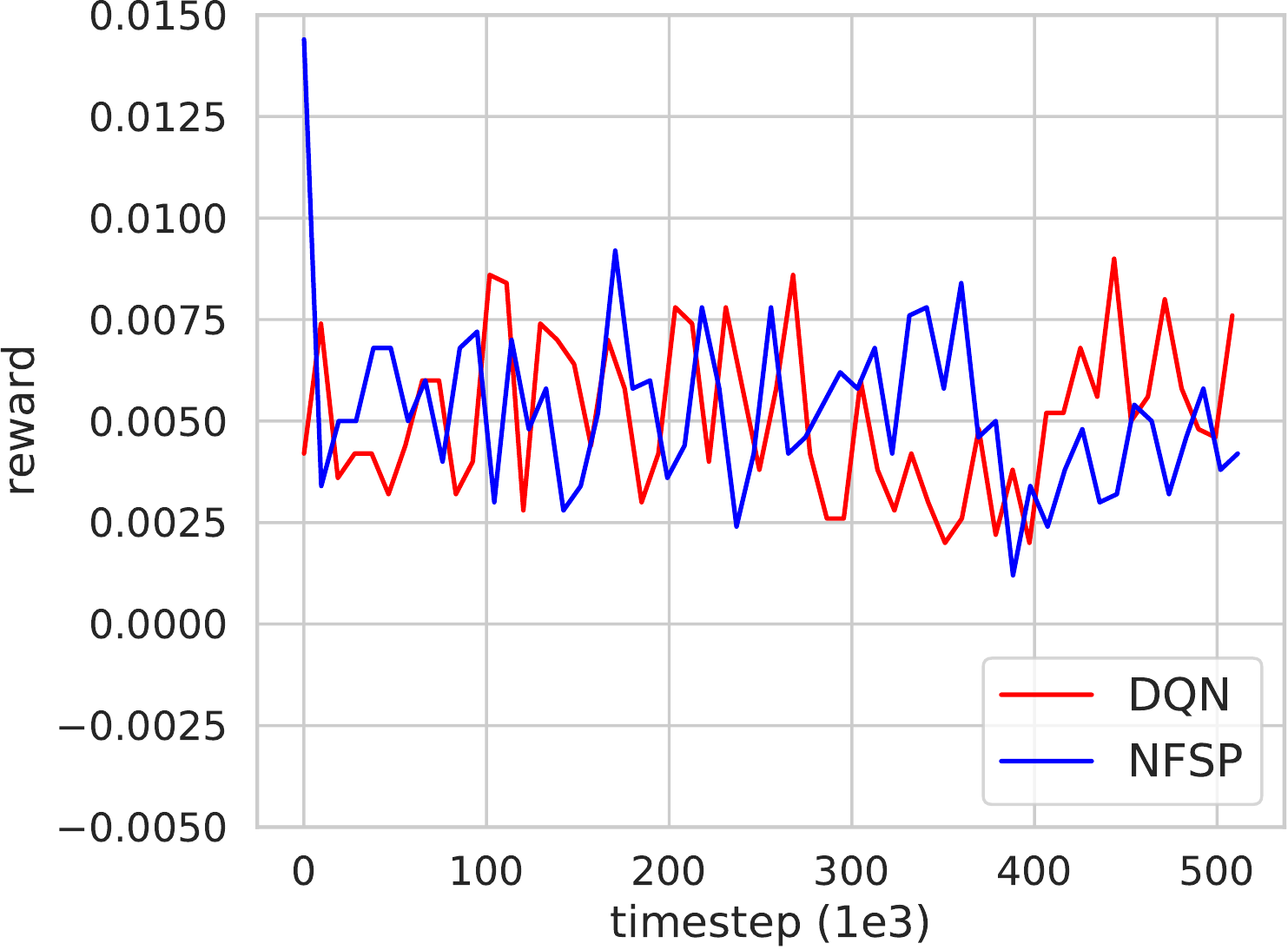}
    \caption{Mahjong}
  \end{subfigure}%
  \begin{subfigure}[b]{0.24\textwidth}
    \includegraphics[width=\textwidth]{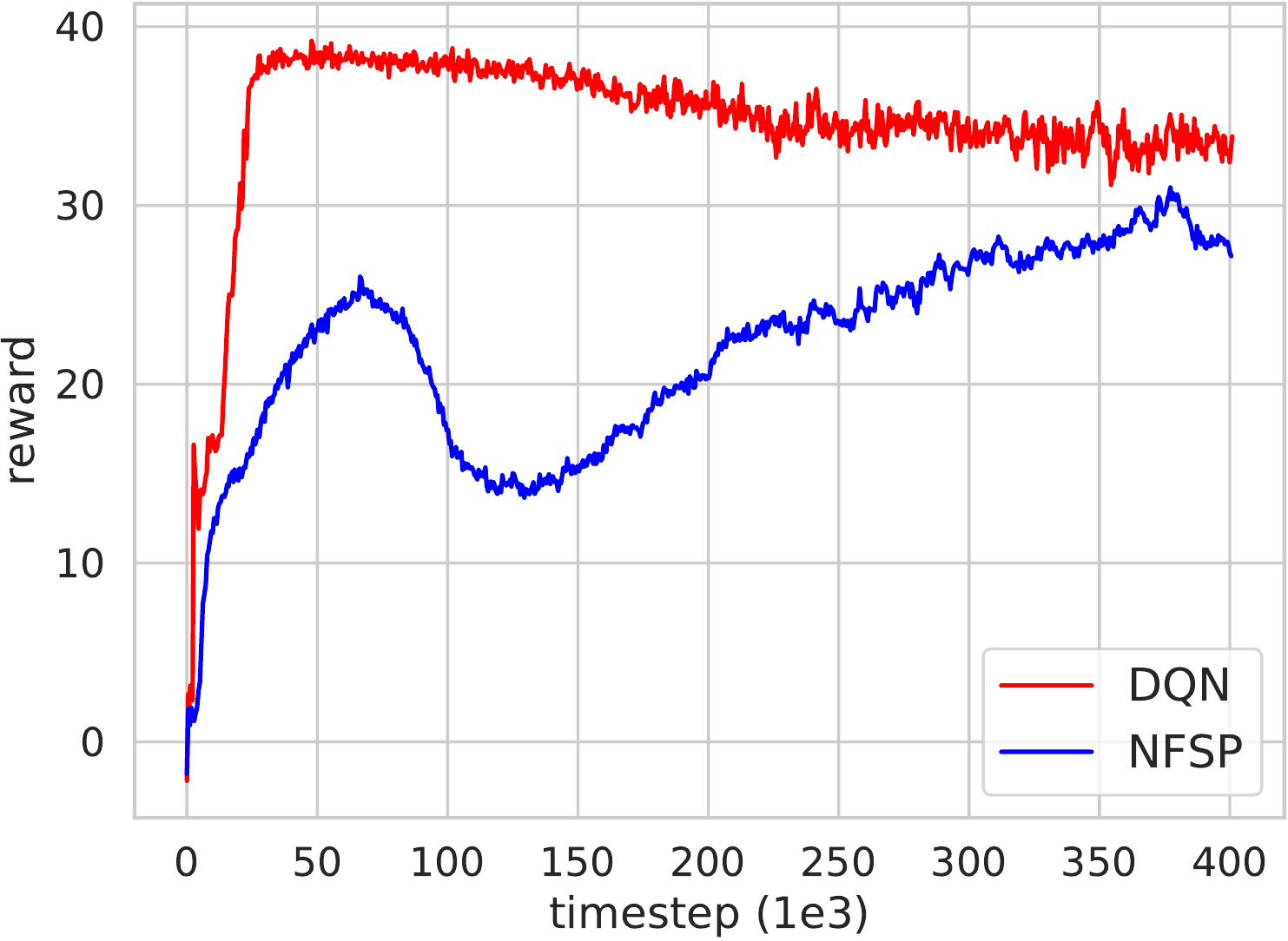}
    \caption{No-limit Texas Hold'em}
  \end{subfigure}%
  \begin{subfigure}[b]{0.24\textwidth}
    \includegraphics[width=\textwidth]{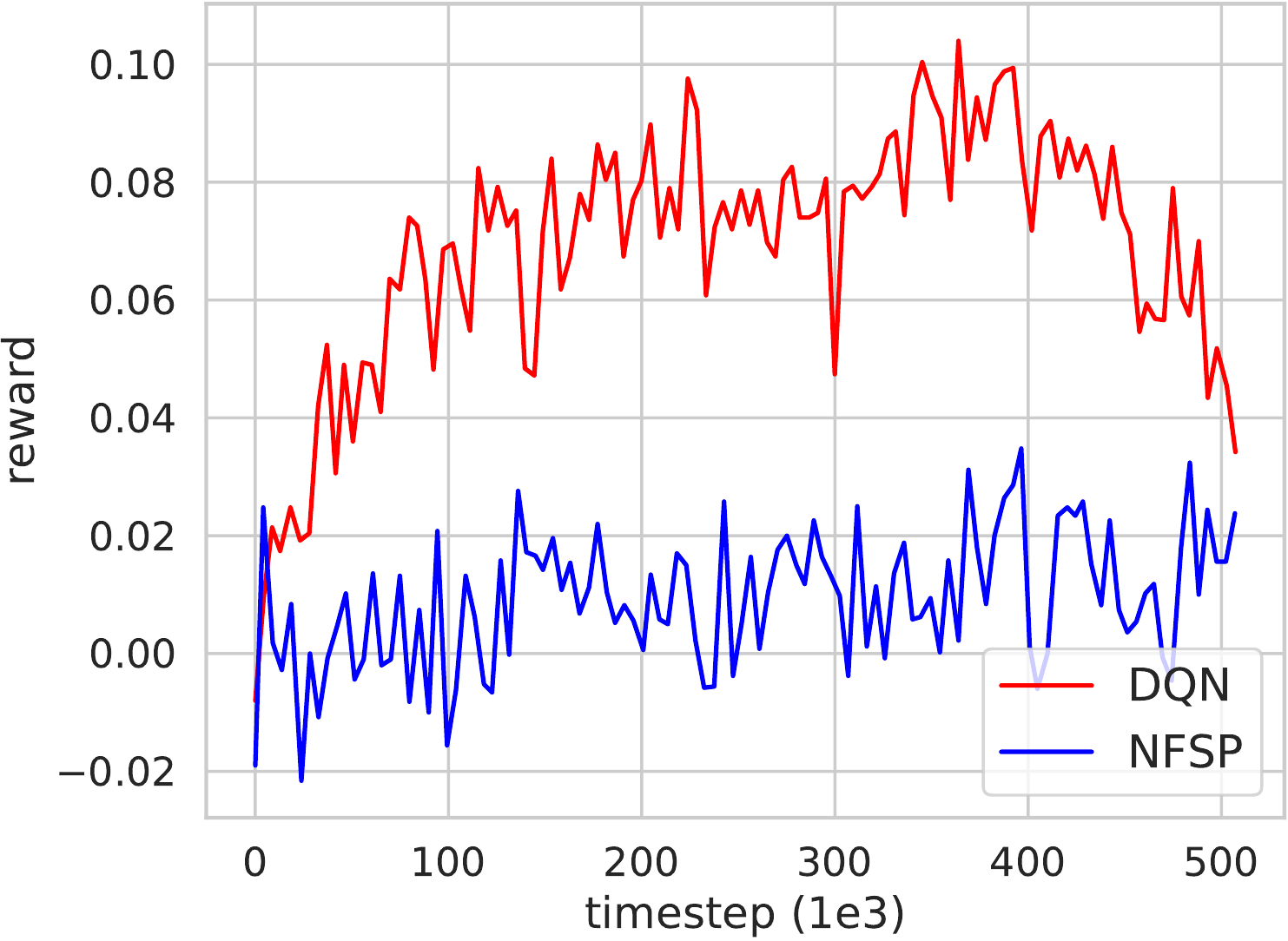}
    \caption{UNO}
  \end{subfigure}%
  
  \caption{Learning curves on the card environments in terms of the performance against random agents. X-axis represents the total steps taken in the environment. Y-axis is the reward achieved by playing against random agents.}
  \label{fig:curve}
\end{figure*}

\subsection{Advanced Interfaces}
We also provide some advanced interfaces that operate upon the game tree. Following other RL toolkits~\cite{brockman2016openai,Marc2019openspiel}, we define a \texttt{step} function which moves the environment to the next state given the current action. To enable traversing backward, we provide a \texttt{step\_back} function, which traverses back to the previous state. Note that the action of the current player will lead to the observation of the next player, and the current player can observe her next state only after all other players have chosen their actions. Thus, users need to be careful with \texttt{step} and \texttt{step\_back} since the next state is ``delayed''.

Note that the deign of \texttt{step} and \texttt{step\_back} is similar to traditional tree-based interface. Specifically, \texttt{step} is corresponding to accessing child node, and \texttt{step} would access the parent node. This design enables flexible node visiting strategies of the game tree, such as external sampling MCCFR~\cite{lanctot2009monte}.



\section{Evaluation}
This section conducts experiments to evaluate the toolkit. We mainly focus on the following two questions: (1) How the state-of-the-art reinforcement learning algorithms perform on the introduced environments? (2) How many computation resources are required to generate game data?

\subsection{Training Agents on Environments}
We apply Deep Q-Network~(DQN)~\cite{silver2016mastering}, Neural Fictitious Self-Play~(NFSP)~\cite{heinrich2016deep}, and Counterfactual Regret Minimization~(CFR)~\cite{zinkevich2008regret} to the environments. These algorithms belong to different categories. DQN is a standard single-agent reinforcement learning algorithm, NFSP is a deep reinforcement learning approach for extensive games with imperfect information, and CFR is a standard regret minimization method for extensive imperfect information games. For DQN, we fix other players as random agents so that DQN agent can be trained in single-agent setting. We only test CFR on Leduc Hold'em since it is computationally expensive, requiring complete traversal of the game tree. Blackjack is only tested by DQN because it is a single-agent environment.

To evaluate the agents is non-trivial. The performance of imperfect information game is usually measured by exploitablity~\cite{zinkevich2008regret,johanson2011accelerating}, which searches for the best response against the trained policy. However, it is computationally expensive to obtain the best response for the large environments in the toolkit since it relies on traversal of the game tree. Thus, we evaluate the performance based on winning rates. In this paper, we adopt two methods to empirically evaluate the performance. First, we report the winning rates of the agents against random agents. Second, we compare the agents with tournaments.
\begin{table*}[t]
    \small
    \centering
    \begin{tabular}{l|C{1.2cm}|C{1.4cm}|C{1.2cm}|C{1.4cm}|C{1.2cm}|C{1.4cm}|C{1.2cm}|c}
         \hline  
         \multirow{2}{*}{} & \multicolumn{2}{c|}{$\times 1$} & \multicolumn{2}{c|}{$\times 4$} & \multicolumn{2}{c|}{$\times 8$} & \multicolumn{2}{c}{$\times 16$} \\
         \cline{2-9}
         & total & per step & total & per step & total & per step & total & per step\\
         \hline
         Blackjack & $156.1$ & $1.1 \times 10^{-4}$ & $45.6$ & $3.3 \times 10^{-5}$ & $23.0$ & $1.7 \times 10^{-5}$ & $12.8$ & $9.3 \times 10^{-6}$\\
         Leduc Hold'em & $204.6$ & $8.0 \times 10^{-5}$ & $58.7$ & $2.3 \times 10^{-5}$ & $28.7$ & $1.1 \times 10^{-5}$ & $17.9$ & $7.0 \times 10^{-6}$ \\
         Limit Texas Hold'em & $324.6$ & $1.1 \times 10^{-4}$ & $83.3$ & $2.8 \times 10^{-5}$ & $45.9$ & $1.6 \times 10^{-5}$ & $24.6$ & $8.3 \times 10^{-6}$ \\
         Dou Dizhu & $87270.8$ & $1.4 \times 10^{-3}$ & $22894.3$ & $3.6 \times 10^{-4}$ & $12753.9$ & $2.0 \times 10^{-4}$ & $7275.9$ & $1.1 \times 10^{-4}$ \\
         Mahjong & $74786.0$ & $8.1 \times 10^{-4}$ & $20825.9$ & $2.3 \times 10^{-4}$ & $11059.7$ & $1.2 \times 10^{-4}$ & $6169.6$ & $6.7 \times 10^{-5}$ \\
         No-limit Texas Hold'em & $597.7$ & $1.4 \times 10^{-4}$ & $160.6$ & $3.7 \times 10^{-5}$ & $81.9$ & $1.9 \times 10^{-5}$ & $48.4$ & $1.1 \times 10^{-5}$ \\
         UNO & $4952.9$ & $1.1 \times 10^{-4}$ & $1366.0$ & $3.0 \times 10^{-5}$ & $696.5$ & $1.5 \times 10^{-5}$ & $440.7$ & $9.5 \times 10^{-6}$ \\

         \hline 
    \end{tabular}
    \vspace{-3pt}
    \caption{Running time in seconds of $1,000,000$ games with random agents, under one process and multiple processes. Per step: the running time divided by the number of performed timesteps.}
    \label{tab:efficiency}
        \vspace{-13pt}
\end{table*}
In our experiments, the hyperparameters are lightly tuned. For DQN, the memory size is selected in $\{2000, 100000\}$, the discount factor is set to $0.99$, Adam optimizer is applied with learning rate $0.00005$, and the network structure is MLP with size 10-10 128-128, 512-512 or 512-1024-2048-1024-512 based on the size of the state and action space. For NFSP, the anticipatory parameter is chosen from $\{0.1, 0.5\}$. Memory size for supervised learning and reinforcement learning are $10^6$ and $3 \times 10^4$, respectively.

\begin{table}[t]
    \centering
        \begin{tabular}{l|c c}
        \hline  
         \textbf{Tournament} & \textbf{NFSP} & \textbf{DQN}\\
         \hline 
         Leduc Hold'em & 1.0691 & -1.0691\\
         Limit Texas Hold'em & -0.0308 & 0.0308\\
         Dou Dizhu with NFSP landlord & 0.7049 & 0.2951\\
         Dou Dizhu with DQN landlord & 0.7303 & 0.2697\\
         Mahjong & -0.0090 & -0.0104 \\
         No-limit Texas Hold'em & 9.5610 & -9.5610\\
         UNO & -0.0428 & 0.0428 \\
         \hline
        
        \end{tabular}
    \caption{Average payoffs of NFSP and DQN by playing $10,000$ games. For Dou Dizhu, we switch roles of landlord and peasants and report the results separately. For Mahjong, two DQN agents and two NFSP agents randomly choose seats in each game, and the averaged results are reported.}
    \label{tab:tournament}
        \vspace{-18pt}
\end{table}

\paratitle{Results against random agents.} The learning curves in terms of the performance against random agents are shown in Figure~\ref{fig:curve}. The rewards of the agents are obtained through competitions against random agents. Specifically, the rewards of betting games~(Leduc Hold'em, Limit Texas Hold'em, No-limit Texas Holdem) are defined as the average winning big blinds per hand. The rewards of the other games are obtained directly from the winning rates. In Dou Dizhu, players are in different roles~(landlord and peasants). We fix the role of the agent as the landlord in evaluation.

We make two observations. First, all the algorithms have similar results against random agents. DQN is slightly better than NFSP on Texas Hold'em and UNO, while NFSP is slightly better than DQN on Leduc Hold'em and Dou Dizhu. It is reasonable for DQN to achieve this result since DQN is trained to exploit the random agents. Second, NFSP and DQN are highly unstable in large games. Specifically, they only achieve minor improvements during the learning process on UNO, Mahjong and Dou Dizhu. These games are challenging due to their large state/action space and sparse reward. We believe there is a lot of room for improvement. More efforts are needed to study how we can stably train reinforcement learning agents in these large environments. 

\paratitle{Tournament results.} We report the average payoffs the agents achieved when playing against each other. The results between NFSP and DQN are shown in Table~\ref{tab:tournament}. We observe that NFSP is stronger than DQN on most of the environments. We also compare CFR with NFSP and DQN on Leduc Hold'em. CRF achieves better performance, winning $0.0776$ and $1.2493$ against NFSP and DQN, respectively.
\vspace{-3pt}

\paratitle{Discussion} We further analyze the trained agents. We find that the DQN agents play very aggressively in betting games. For example, in Leduc Hold'em environment, DQN agent tend to choose ``raise'' or ``call'' in almost every decision. Interestingly, this naive strategy works well when the opponent is a random agent since the random agent may easily choose ``fold'' so that that the DQN policy can win. However, DQN policy may be highly exploitable since one can easily find its weaknesses. Thus, performance against random agents can only be a way to get a sense of whether the agent is improving on the environment, but is \textbf{NOT} enough to be used to evaluate algorithms. For large games, we recommend evaluating algorithms by playing against existing models. To benchmark the evaluation, we will develop rule-based agents and stronger pre-trained models in the future.

\subsection{Running Time Analysis}
We evaluate the efficiency of the implemented environments by running self-play on the games with random agents. Specifically, we report the running time in seconds of $1,000,000$ games using a single process and multiple processes. Since Dou Dizhu, UNO and Mahjong have long sequences in one game, we additionally report a normalized version of running time, i.e., the average running time per timestep. Our experiments are conducted on a server with 24 Intel(R) Xeon(R) Silver 4116 CPU @2.10GHz processors and 64.0 GB memory. Each experiment is run $3$ times with different random seeds. The average running time in seconds is reported in Table~\ref{tab:efficiency}. We observe that all the environments achieve higher throughputs with more processors.

\section{Related work}
There are a few open-source reinforcement learning libraries, most of which focus on single-agent environments~\cite{brockman2016openai,duan2016benchmarking,shi2019virtual}. Recently, there have been some projects that support multi-agent environments~\cite{vinyals2017starcraft,zheng2018magent,juliani2018unity,suarez2019neural}. However, they do not support card game environments. A contemporary framework OpenSpiel~\cite{Marc2019openspiel} provides a large collection of games, including several simple card games. Our toolkit is specifically designed for card games with straightforward interfaces, supporting various styles of card games that are not included in existing toolkits. 

The most popular techniques for solving poker games in literature are Counterfactual Regret Minimization~(CFR)~\cite{zinkevich2008regret} and its variants~\cite{brown2018deep}. Achievements have been made on betting games such as Texas Hold'em~\cite{moravvcik2017deepstack,brown2017safe}. However, CFR is computationally expensive, since it relies on complete traversal of the game tree, and is infeasible for games with large state space such as Dou Dizhu~\cite{deltadou}.

Recent studies show that reinforcement learning strategies can perform well in betting games~\cite{heinrich2015fictitious,heinrich2016deep,lanctot2017unified}, and achieve satisfactory performance in Dou Dizhu~\cite{deltadou}. The inspiring results and the flexibility of RL offer the opportunity to explore deep reinforcement learning in more difficult card games with large state and action space.
\vspace{-5pt}
\section{Conclusions and Future Directions}
In this paper, we introduce RLCard, an open-source toolkit for reinforcement learning research in card games. RLCard supports multiple challenging card environments wrapped with common and easy-to-use interfaces. In the future, we plan to enhance the toolkit in several aspects. First, in order to benchmark the evaluation, we would like to design rule-based agents and provide more pre-trained models for evaluation. Second, we plan to develop visualization and analysis tools for the environments. Third, we will further accelerate the environments with more efficient implementations. Last, we will include more interesting games and more algorithms to enrich the toolkit.

\section{Acknowledgements}
We would like to thank JJ World Network Technology Co., LTD for the generous support.
\vspace{-6.5pt}













\bibliographystyle{aaai}
\bibliography{reference}

\appendix
\section{Appendix}

\subsection{Game Design}
\label{sec:2:2}
Card games are usually played following similar procedures. We design several abstract base classes which are implemented in the specific games. In the toolkit, some common concepts are abstracted and defined as follows:
\begin{itemize}
    \item \textbf{Player.} The person who plays the game. Each game is usually played by multiple players.
	\item \textbf{Game.} A game is a complete sequence starting from one of the non-terminal states to a terminal state. At the end of a game, each player will receive a payoff.
	\item \textbf{Round.} A round is a part of the sequence of a game. Most card games can be naturally divided into multiple rounds. For instance, Texas Hold'em consists of four rounds of betting. In Doudi Zhu a round is finished when two consecutive players pass.
	\item \textbf{Dealer.} Card games usually require shuffling and allocating a deck of cards to players. Dealer is responsible for deck management.
	\item \textbf{Judger.} A judger is responsible for making major decisions in a round or at the end of a game. For example, the next player in UNO is decided based on the type of the last card. In Texas Hold'em, the payoff is determined in the end of a game.
\end{itemize}
The games in the toolkit are implemented by associating a class with each of the above concepts. The common design principle makes the game logic easy to follow and understand. Other card games are usually compatible with the above structure so that can be easily added to the toolkit.

\vspace{-4pt}
\subsection{Environment Interfaces}
This section briefly introduces the interfaces of the toolkit. We describe state representation, action encoding, and how we can modify them to customize the environments. After that, we show how to generate data with multiple processes. Finally, we introduce a single-agent interface, where the other players are simulated by pre-trained or rule-based models. More documents and examples can be found at the Github repository.

\vspace{-4pt}
\paratitle{State Representation}
State is defined as all the information that can be observed from the view of one player in a specific timestep of the game. In the toolkit, each state is a dictionary consisting of two values. The first value is a list of legal actions. The second value is observation. There are various ways to encode the observation. For Blackjack, we directly use the player's score and the dealer's score as a representation. For other games in the toolkit, we encode the observed cards into several card planes. For example, in Dou Dizhu, the input of the policy is a matrix of $6$ card planes, including the current hand, the union of the other two players' hands, the recent three actions, and the union of all the played cards.

\vspace{-4pt}
\paratitle{Action Encoding}
The specific actions in a game are all encoded into action indices, which are positive integers starting from $0$. Each action index corresponds to exactly one action in the game. The legal actions are also represented as a list of action indices. At each step, an agent should choose one of the action indices (i.e., integer values) among the legal actions instead of specific actions (such as ``hit'' or ``stand'' in Blackjack).

For some large games, action abstraction is adopted to reduce the action space. For example, Dou Dizhu suffers from the combinatorial explosion of the action space with more than $3 \times 10^4$ actions, where any trio, plane or quad can be combined with any individual card or pair (kicker). To reduce the action space, we only encode the major part of a combination and use rules to decide the kicker. In this way, the action space of Dou Dizhu is reduced to $309$.

\vspace{-4pt}
\paratitle{Customization}
In addition to the default state and action encoding, our design enables customization of state representation, action encoding, reward design, and even the game rules.

Each game is wrapped by an \texttt{Env} class, in which we can rewrite some key functions to customize the environments. The function of \texttt{extract\_state} is to convert the original game state into representation. The function of \texttt{decode\_action} is to map action indices to actions. One can implement his own abstraction of the actions by modifying this function. The function of \texttt{get\_payoffs} will return the payoffs of the players in the end of the game. For each game, we provide a default setting for each of the above components. Users are encouraged to customize these settings to achieve better performance.

The parameters of each game can also be adjusted. For example, one can change the number of players or the fixed raise in Limit Texas Hold'em by modifying \texttt{\_\_init\_\_} function of the \texttt{LimitholdemGame} class. In this way, we may adjust the difficulty of the games for our purposes and design algorithms step by step.

\vspace{-4pt}
\paratitle{Parallel Training}
The toolkit supports generating game data with multiple processes. Running in parallel will greatly accelerate the training in large environments. Specifically, we create duplicate environments in initialization. Then each worker will copy the model parameters from the main process, generate game data in a duplicate of the environment, and send the data to the main process. The main process will collect all the data to train the agent on either CPU or GPU. Example implementations of training agents with multiple processes can be found at the Github repository.

\vspace{-4pt}
\paratitle{Single-Agent Interfaces}
We provide interfaces to explore training single-agent reinforcement learning agents in card games. Specifically, we develop pre-trained or rule-based models to simulate other players so that the games essentially become single-agent environments from the view of one player. These single-agent environments are also challenging since they have large state and action space, and sparse reward. In the future, we plan to use different levels of simulating models to create environments with various difficulties. The single-agent interfaces follow OpenAI Gym~\cite{brockman2016openai}. Specifically, in the single-agent mode, given an action, \texttt{step} function will return the next state, reward, and whether the game is done. The \texttt{reset} function will reset the game and return the initial state. Standard single-agent RL algorithms can be easily applied to the environments.


\end{document}